\documentclass[11pt,a4paper]{article}
\pdfoutput=1
\usepackage{emnlp-ijcnlp-2019}
\usepackage{times}
\usepackage{latexsym}
\usepackage{url}
\usepackage{amsmath}
\usepackage{amsthm}
\usepackage{amssymb}
\usepackage{dsfont}
\usepackage{multirow}
\usepackage{bm}
\usepackage{mathtools}
\usepackage{xspace}
\usepackage[section]{algorithm}
\usepackage[noend]{algpseudocode}
\usepackage{graphicx}
\usepackage{xr}
\usepackage {diagbox}

\aclfinalcopy 

\makeatletter
\DeclareRobustCommand\onedot{\futurelet\@let@token\@onedot}
\def\@onedot{\ifx\@let@token.\else.\null\fi\xspace}
\def\eg{\emph{e.g}\onedot} 
\def\ie{\emph{i.e}\onedot}

\makeatother

\newcommand{\Tref}[1]{Table~\ref{#1}}
\newcommand{\Eref}[1]{Eq.~(\ref{#1})}
\newcommand{\Fref}[1]{Fig.~\ref{#1}}
\newcommand{\Sref}[1]{Sec.~\ref{#1}}
\newcommand{\Aref}[1]{Algorithm~\ref{#1}}
\newcommand{\Lref}[1]{line~\ref{#1}}

\algrenewcommand\algorithmicindent{0.7em}%

\title{Transfer Fine-Tuning: A BERT Case Study}

\author{Yuki Arase$^{1 \star}$ \and Junichi Tsujii$^{\star 2}$ \\
$^1$Osaka University, Japan\\
$^\star$Artificial Intelligence Research Center (AIRC), AIST, Japan\\
$^2$NaCTeM, School of Computer Science, University of Manchester, UK \\
 {\tt arase@ist.osaka-u.ac.jp},  {\tt j-tsujii@aist.go.jp}\\
 }

\date{}

\begin{document}
\maketitle
\begin{abstract}
A semantic equivalence assessment is defined as a task that assesses semantic equivalence in a sentence pair by binary judgment (\ie, paraphrase identification) or grading (\ie, semantic textual similarity measurement). 
It constitutes a set of tasks crucial for research on natural language understanding. 
Recently, BERT realized a breakthrough in sentence representation learning~\cite{bert}, which is broadly transferable to various NLP tasks. 
While BERT's performance improves by increasing its model size, the required computational power is an obstacle preventing practical applications from adopting the technology. 
Herein, we propose to inject phrasal paraphrase relations into BERT in order to generate suitable representations for semantic equivalence assessment instead of increasing the model size. 
Experiments on standard natural language understanding tasks confirm that our method effectively improves a smaller BERT model while maintaining the model size. 
The generated model exhibits superior performance compared to a larger BERT model on semantic equivalence assessment tasks. 
Furthermore, it achieves larger performance gains on tasks with limited training datasets for fine-tuning, which is a property desirable for transfer learning.  
\end{abstract}

\section{Introduction}
Paraphrase identification and semantic textual similarity (STS) measurements aim to assess semantic equivalence in sentence pairs. 
These tasks are central problems in natural language understanding research and its applications. 
In this paper, these tasks are defined as semantic equivalence assessments.

Sentence representation learning is the basis of assessing semantic equivalence. 
Unsupervised learning is becoming the preferred approach because it only requires plain corpora, which are now abundantly available. 
In this approach, a model is pre-trained to generate generic sentence representations that are broadly transferable to various natural language processing (NLP) tasks. 
Subsequently, it is fine-tuned to generate specific representations for solving a target task using an annotated corpus. 
Considering the high costs of annotation, a pre-trained model that efficiently fits the target task with a smaller amount of annotated corpus is desired.

Recently, Bidirectional Encoder Representations from Transformers (BERT) realized a breakthrough, which dramatically improved sentence representation learning~\cite{bert}.
BERT pre-trains its encoder using language modeling and by discriminating surrounding sentences in a document from random ones. 
Pre-training in this manner allows distributional relations between sentences to be learned. 
Intensive efforts are currently being made to pre-train larger models by feeding them enormous corpora for improvement~\cite{opeai-gpt2,xlnet}. 
For example, a large model of BERT has $340$M parameters, which is $3.1$ times larger than its smaller alternative. 
Although such a large model achieves performance gains, the required computational power hinders its application to downstream tasks.

Given the importance of natural language understanding research, we focus on sentence representation learning for semantic equivalence assessment. 
Instead of increasing the model size, we propose the injection of semantic relations into a pre-trained model, namely BERT, to improve performance. 
\newcite{phang-arxiv-2018} showed that BERT's performance on downstream tasks improves by simply inserting extra training on data-rich supervised tasks. 
Unlike them, we inject semantic relations of finer granularity using phrasal paraphrase alignments automatically identified by~\newcite{arase:emnlp2017} to improve semantic equivalent assessment tasks. 
Specifically, our method learns to discriminate phrasal and sentential paraphrases on top of the representations generated by BERT.  
This approach explicitly introduces the concept of the phrase to BERT and supervises semantic relations between phrases. 
Due to studies on sentential paraphrase collection~\cite{D17-1126} and generation~\cite{P18-1042}, a million-scale paraphrase corpus is ready for use.  
We empirically show that further training of a pre-trained model on relevant tasks transfers well to downstream tasks of the same kind, which we name as {\em transfer fine-tuning}.

The contributions of our paper are:
{
\setlength{\leftmargini}{15pt}
\begin{itemize}
\setlength{\itemsep}{0pt}
\item We empirically demonstrate that transfer fine-tuning using paraphrasal relations allows a smaller BERT to generate representations suitable for semantic equivalence assessment. The generated model exhibits superior performance to the larger BERT while maintaining the small model size. 
\item Our experiments indicate that phrasal paraphrase discrimination contributes to representation learning, which complements simpler sentence-level paraphrase discrimination. 
\item Our model exhibits a larger performance gain over the BERT model for a limited amount of fine-tuning data, which is an important property of transfer learning. 
\end{itemize}
}
\noindent
We hope that this study will open up one of the crucial research directions that will make the approach of pre-trained models more practically useful.
Our codes, datasets, and the trained models will be made publicly available at our web site. 

  
\section{Related Work}
Sentence representation learning is an active research area due to its importance in various downstream tasks. 
Early studies employed supervised learning where a sentence representation is learned in an end-to-end manner using an annotated corpus. 
Among these, the importance of phrase structures in representation learning has been discussed~\cite{tai-socher-manning:2015:ACL-IJCNLP,D18-1408}. 
In this paper, we use structural relations in sentence pairs for sentence representations. 
Specifically, we employ phrasal paraphrase relations that introduce the notion of a phrase to the model.

The research focus of sentence representation learning has moved toward unsupervised learning in order to exploit the gigantic corpus. 
Skip-Thought, which was an early learning attempt, learns to generate surrounding sentences given a sentence in a document~\cite{NIPS2015_5950}.  
This can be interpreted as an extension of the distributional hypothesis on sentences. 
Quick-Thoughts, a successor of Skip-Thought, conducts classification to discriminate surrounding sentences instead of generation~\cite{logeswaran2018an}. 
GenSen combines these approaches in massive multi-task learning~\cite{subramanian2018learning} based on the premise that learning dependent tasks enriches sentence representations.

Embeddings from Language Models (ELMo) made a significant step forward~\cite{elmo}. 
ELMo uses language modeling with bidirectional recurrent neural networks (RNN) to improve word embeddings. 
ELMo's embedding contributes to the performance of various downstream tasks. 
OpenAI GPT~\cite{opeai-gpt} replaced ELMo's bidirectional RNN for language modeling with the Transformer~\cite{transformer} decoder.  
More recently, BERT combined the approaches of Quick-Thoughts (\ie, a next-sentence prediction approach) and language modeling on top of the deep bidirectional Transformer. 
BERT broke the records of the previous state-of-the-art methods in eleven different NLP tasks. 
While BERT's pre-training generates generic representations that are broadly transferable to various NLP tasks, we aim to fit them for semantic equivalence assessment by injecting paraphrasal relations. 
\newcite{bigbird} showed that BERT's performance improves when fine-tuning with a multi-task learning setting, which is applicable to our trained model for further improvement.

\section{Background}
\subsection{Phrase Alignment for Paraphrases}
\label{sec:phrase_alignment}
\begin{figure}[!t]
\centering
\includegraphics[width=0.9\linewidth]{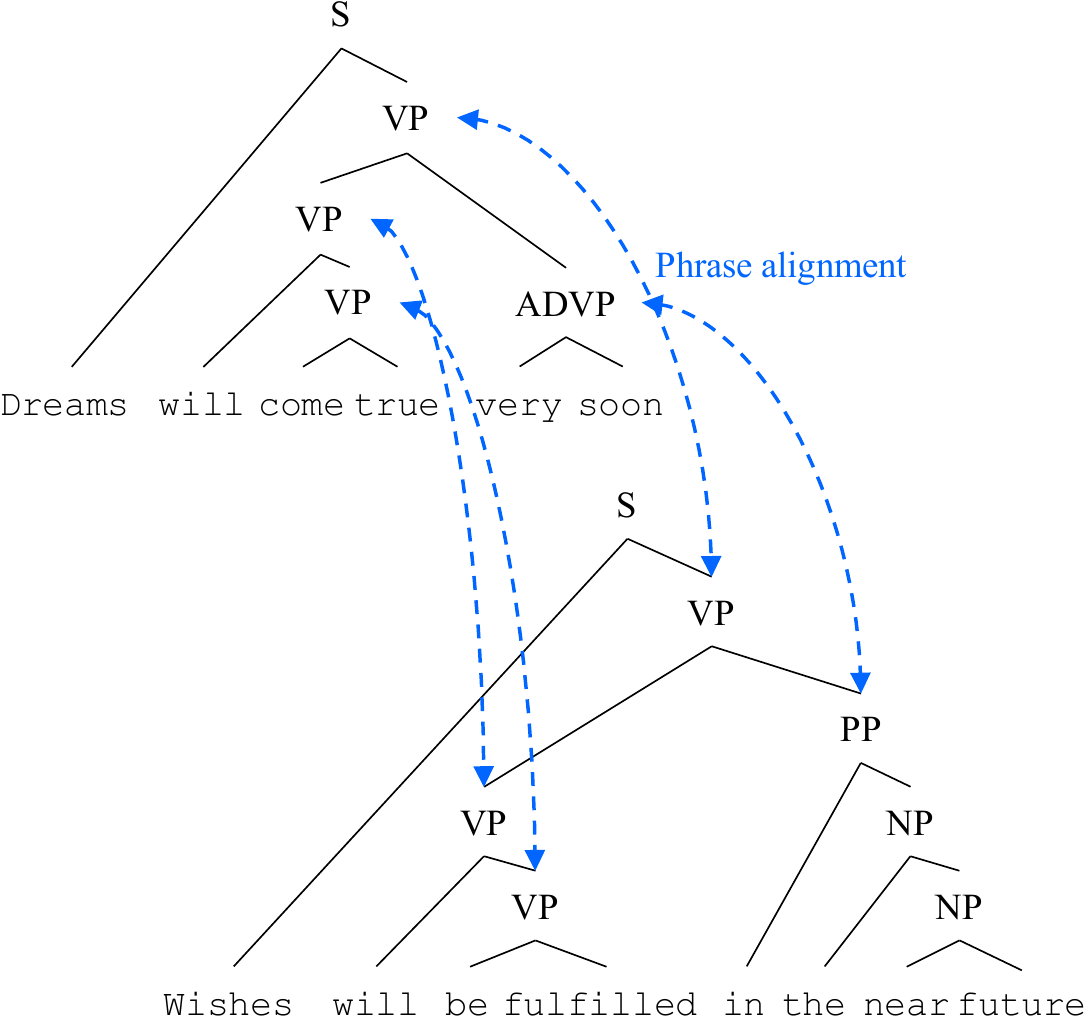}
\caption{Phrasal paraphrases are obtained from~\cite{arase:emnlp2017}; arrows indicate phrase alignments.}
\label{fig:phrase_alignment}
\end{figure} 
In order to obtain phrasal paraphrases, we used the phrase alignment method proposed in~\cite{arase:emnlp2017} and apply it to our paraphrase corpora. 
The alignment method aligns phrasal paraphrases on the parse forests of a sentential paraphrase pair as illustrated in \Fref{fig:phrase_alignment}.

According to the evaluation results reported in~\cite{arase:emnlp2017}, the precision and recall of alignments are $83.6\%$ and $78.9\%$, which are $89\%$ and $92\%$ of those of humans, respectively. 
Although alignment errors occur, previous studies show that neural networks are relatively robust against noise in a training corpus and still benefit from extra supervisions as demonstrated in~\cite{D18-1045,P18-1080}.

We collect all the spans of phrases in a sentential paraphrase pair and their alignments as pairs of phrase spans. 
Because the phrase alignment method allows unaligned phrases, not all of the phrases have aligned counterparts. 

\subsection{Pre-Training on BERT}
\label{sec:bert}
BERT is a bidirectional Transformer that generates a sentence representation by conditioning both the left and right contexts of a sentence. 
A pre-trained BERT model can be easily fine-tuned for a wide range of tasks by just adding a fully-connected layer, without any task-specific architectural modifications. 
BERT achieved state-of-the-art performances for eleven NLP tasks, thereby outperforming the previous state-of-the-art methods by a large margin.

Pre-training in BERT accomplishes two tasks. 
The first task is masked language modeling, where some words in a sentence are randomly masked and the model then predicts them from the context. 
This task design allows the representation to fuse both the left and the right context. 
The second task predicts whether a pair of sentences are consecutive in a document to learn the relation between the sentences.  
Specifically, as illustrated in \Fref{fig:overview}, BERT takes two sentences as input that are concatenated by a special token {\tt [SEP]}.\footnote{Throughout the paper, {\tt typewriter font} represents tokens and labels.} 
The first token of every input is always the special token of {\tt [CLS]}. 
The final hidden state corresponding to this {\tt [CLS]} token is regarded as an aggregated representation of the input sentence pair. 
This is used to predict whether the sentence pair is composed of consecutive sentences in a document or not during pre-training.

BERT has a deep architecture. 
The BERT-base model has $12$ layers of $768$ hidden size and $12$ self-attention heads. 
The BERT-large model has $24$ layers of $1024$ hidden size and $16$ self-attention heads. 
Both BERT-bese and BERT-large models were pre-trained using BookCorpus~\cite{Zhu_2015_ICCV} and English Wikipedia (in total $3.3$B words).

\section{Transfer Fine-Tuning with Paraphrasal Relation Injection}
\label{sec:our-model}
We inject semantic relations between a sentence pair into a pre-trained BERT model through classification of phrasal and sentential paraphrases. 
After the training, the model can be fine-tuned in exactly the same manner as with BERT models.   

\begin{algorithm} [!t]
\caption{Paraphrasal Relation Injection}
\label{alg:tailoring}
\begin{algorithmic}[1] 
\Require Paraphrase sentence pairs $P=\{\langle s, t \rangle\}$ , a pre-trained BERT model
\State Obtain a set of phrase alignments $A$ as pairs of spans for each $\langle s, t \rangle \in P$ \label{ln:phrase_alignment}
\State WordPiece tokenization of $P$ \label{ln:tokenization}
\State Accommodate phrase spans in $A$ to BERT's token indexing: $A=\{\langle (j, k),  (m, n) \rangle\}$ \label{ln:span_adjustment}
\Repeat
\ForAll{mini-batch  $b_t \in \{\langle P_i, A_i \rangle\}$}
\State Encode $b_t$ by the BERT model
 \State Compute loss: L($\Theta$)
 \State \hspace{\algorithmicindent} For phrasal paraphrase task: $L_p(\Theta)$ \label{ln:phrase_classification}
 \State \hspace{\algorithmicindent} For sentential paraphrase task: $L_s(\Theta)$ \label{ln:sentence_classification}
 \State \hspace{\algorithmicindent} $L(\Theta)=L_p(\Theta)+L_s(\Theta)$
 \State Compute gradient: $\nabla(\Theta)$
 \State Update the model parameters
\EndFor
\Until{convergence}
\end{algorithmic} 
\end{algorithm} 

\subsection{Overview}
\begin{figure*}[!t]
\centering
\includegraphics[width=0.98\linewidth]{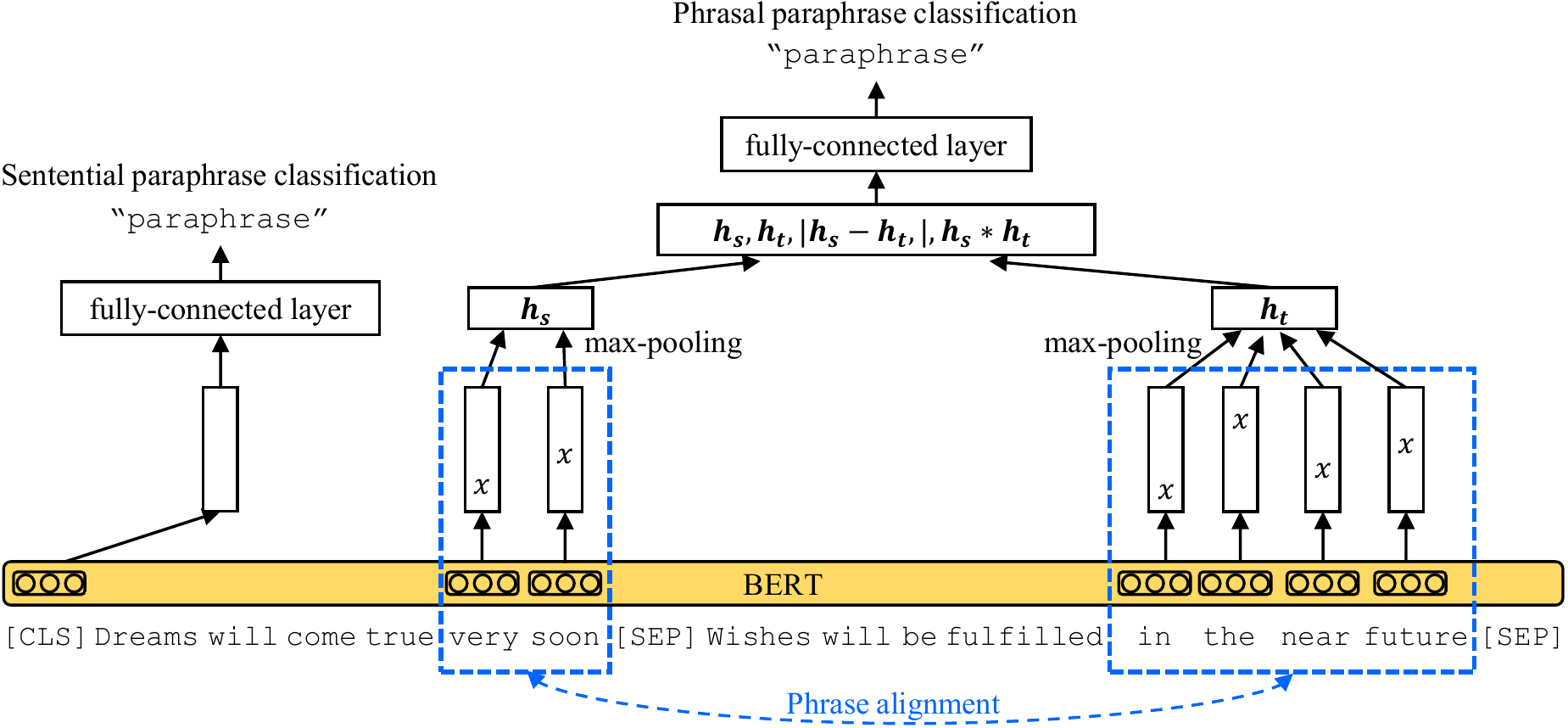}
\caption{Our method injects semantic relations to sentence representations through paraphrase discrimination.}
\label{fig:overview}
\end{figure*} 

\Aref{alg:tailoring} provides an overview of our method. 
It takes a sentential paraphrase pair $\langle s, t \rangle$ as an input, which are referred to as the source and target, respectively, for the sake of clarity.  
First, a set of phrase alignments $A$ is obtained for $\langle s, t \rangle$ (\Lref{ln:phrase_alignment}) as described in \Sref{sec:phrase_alignment}. 
Because BERT uses sub-words as a unit instead of words, all the input sentences are tokenized (\Lref{ln:tokenization}) by WordPiece~\cite{wordpiece}. 
In addition, $t$ is concatenated to $s$ when being input to the BERT model, where the first token should always be {\tt [CLS]} and the sentence pair is separated by {\tt[SEP]} as described in \Sref{sec:bert}. 
In order to accommodate to these factors, phrase spans in alignments $A$ are adjusted accordingly (\Lref{ln:span_adjustment}).

Our method learns to discriminate phrasal and sentential paraphrases simultaneously as illustrated in \Fref{fig:overview}. 
Cross-entropy is used as the loss functions for both tasks (\Lref{ln:phrase_classification}, \ref{ln:sentence_classification}).

\paragraph{Phrasal Paraphrase Classification}
The middle part of \Fref{fig:overview} illustrates phrasal paraphrase classification. 
We first generate phrase embedding for each aligned phrase as follows. 
The tokenized sentence pair is encoded by the BERT model. 
For the input sequence of $N$ tokens $\{w_i\}_{i=1,\ldots, N}$, we obtain the final hidden states $\{{\bf h}_i\}_{i=1,\ldots, N}$ (\ie, output of the bidirectional Transformer):
\[
{\bf h}_i=\text{Transformer}(w_1,\ldots,w_N),
\]
where ${\bf h}_i \in \mathbb{R}^\lambda$ and $\lambda$ is the hidden size. 
We then combine $\{{\bf h}_i\}_i$ for a phrase pair with an alignment $\langle (j, k),  (m, n) \rangle$ where $2\leq j < k< m < n \leq N-1$ represent indexes of the beginning and ending of phrases (recall that the first and last tokens are always special tokens in BERT). 
As a combination function, we apply max-pooling that showed strong performance in~\cite{D17-1070} to generate a representation of source and target phrases: 
\begin{eqnarray}
{\bf h}_s&=&\text{max-pooling}({\bf h}_j,\ldots, {\bf h}_k),\label{eq:max-pooling-s}\\
{\bf h}_t&=&\text{max-pooling}({\bf h}_m,\ldots,{\bf h}_n).	\label{eq:max-pooling-t}
\end{eqnarray}
The $\text{max-pooling}(\cdot)$ function selects the maximum value over each dimension of the hidden units.

Then ${\bf h}_s$ and ${\bf h}_t$ are converted to a single vector. 
To extract relations between ${\bf h}_s$ and ${\bf h}_t$, three matching methods are used~\cite{D17-1070}: (a) concatenating the representations $({\bf h}_s, {\bf h}_t)$, (b) taking the element-wise product ${\bf h}_s * {\bf h}_t$, and (c) finding the absolute element-wise difference $|{\bf h}_s-{\bf h}_t|$. 
The final vector of $\mathbb{R}^{4\lambda}$ is fed into a classifier.\footnote{Our follow-up study confirms that a simpler feature generation improves the generality of our model to contribute not only to semantic equivalent assessment but also natural language inference. For details, please refer to the Appendix.}

Because our method aims to generate representations for semantic equivalence assessment,  the classifier should be simple~\cite{logeswaran2018an}. 
Otherwise, a sophisticated classifier would fit itself with the task instead of the representations. 
We use a single fully-connected layer culminating in a softmax layer as our classifier.

Previous studies have calculated interactions between words~\cite{N16-1108} and phrases~\cite{P17-1152} using the final hidden states of bidirectional RNN or recursive neural networks when composing a sentence representation. 
Our approach differs from these by giving explicit supervision of which phrase pairs have semantic interactions (\ie, paraphrases). 

\paragraph{Negative Example Selection}
In paraphrase identification, non-paraphrases with large lexical differences are easy to discriminate. 
Discrimination becomes far more difficult when they contain a number of identical or related words. 
To effectively supervise the model by solving difficult discrimination problems, we designed a three-way classification task: discrimination of {\tt paraphrase},  {\tt random}, and {\tt in-paraphrase} pairs.

The {\tt random} examples are generated by pairing $s$ to a random sentence $t'$ from the training corpus, and then pairing all phrases in $s$ to randomly chosen phrases in $t'$. 
The {\tt in-paraphrase} examples aim to make the discrimination problem difficult, which requires distinguishing true paraphrases and phrases in the paraphrasal sentence pair $t$. 
These may provide sub-phrases or ancestor phrases of true paraphrases as difficult negative examples, which tend to retain the same topic and similar wordings. 
To prepare such examples, for each phrase pair $\langle (j, k),  (m, n) \rangle \in A$, the target span $(m, n)$ is replaced by a randomly chosen phrase span in $t$.

Phrasal paraphrase classification aims to give explicit supervision of semantic relations among phrases in representation learning. 
It also introduces structures in sentences, which is completely missed in BERT's pre-training. 
\newcite{swabha:emnlp2018} showed that supervision of phrase-based syntax improves the performance of a task relevant to semantics, \eg, semantic role labeling. 

\paragraph{Sentential Paraphrase Classification}
The left side of \Fref{fig:overview} illustrates the sentential paraphrase classification. 
The process is simple; the final hidden state of the {\tt [CLS]} token, \ie, ${\bf h}_1$, is fed into a classifier to discriminate whether a sentence pair is a paraphrase or a random sentence combination. 
Note that these random sentence pairs provide {\tt random} phrases for the phrasal paraphrase classification described above. 


\subsection{Training Setting}
We collected paraphrases from various sources as summarized in \Tref{tb:paraphrase_corpus}, which shows the numbers of sentential and phrasal paraphrase pairs after phrase alignment.\footnote{The numbers of sentential paraphrase pairs were reduced due to parsing and alignment failures.}
All the datasets were downloaded from the Linguistic Data Consortium (LDC) or authors' websites. 
The following bullets describe the sources.
{
\setlength{\leftmargini}{15pt}
\begin{itemize}
\setlength{\itemsep}{0pt}
\item NIST OpenMT\footnote{LDC catalogue number: LDC2010T14, LDC2010T17, LDC2010T21, LDC2010T23, LDC2013T03}:~
We randomly paired reference translations of the same source sentence as was done in~\cite{arase:emnlp2017}.
\item Twitter URL corpus~\cite{D17-1126}:~
This corpus was collected from Twitter by linking tweets through shared URLs. We used a three-month collection of paraphrases.\footnote{\url{https://github.com/lanwuwei/Twitter-URL-Corpus}}
\item Simple Wikipedia~\cite{P13-1151}:~
This corpus aligned English Wikipedia and Simple English Wikipedia for text simplification. 
We used ``sentence-aligned, version $2.0$.''\footnote{\url{http://www.cs.pomona.edu/~dkauchak/simplification/data.v2/sentence-aligned.v2.tar.gz}}
\item Para-NMT~\cite{P18-1042}:~
This corpus was created by translating the Czech side of a large Czech-English parallel corpus and pairing the translated English and originally target-side English as paraphrases. 
We used ``Para-nmt-$5$m-processed.''\footnote{\url{https://drive.google.com/file/d/19NQ87gEFYu3zOIp_VNYQZgmnwRuSIyJd/view?usp=sharing}}
\end{itemize}
}

\begin{table}[!t]
\centering
\begin{tabular}{  l | c | c }
\hline
	Source & Sentence & Phrase \\ \hline
	NIST OpenMT & $47$k & $711$k  \\
	Simple Wikipedia & $97$k & $1.4$M \\ 
	Twitter URL corpus & $50$k & $396$k \\
	Para-NMT & $3.9$M & $26.7$M \\ \hline
	Total & $4.1$M & $29.2$M  \\\hline
\end{tabular}
\caption{Numbers of sentential and phrasal paraphrases after the phrase alignment process.}
\label{tb:paraphrase_corpus}
\end{table}
\noindent
Note that these sentential and phrasal paraphrases are obtained by automatic methods. 
On the contrary, dataset creation for downstream tasks generally requires expensive human annotation.

We employed the pre-trained BERT-base model\footnote{\url{https://github.com/google-research/bert}} and conducted paraphrase classification using the collected paraphrase corpora. 
Adam~\cite{adam} was applied as an optimizer with a learning rate of $5\mathrm{ e }-5$. 
A dropout probability was $0.2$ for the fully-connected layers in the classifiers. 
A development set and a test set, each with $50$k sentence pairs, were subtracted from the paraphrase corpus. 
The rest of the corpus was used for training. 
The training was conducted on four NVIDIA Tesla V$100$ GPUs with a batch-size of $100$. 
Early stopping was applied to stop training at the second time decrease in the accuracy of the phrasal paraphrase classification, which was measured on the development set. 
The final test-set accuracies were $98.1\%$ and $99.9\%$ for phrasal and sentential paraphrase classification, respectively.

\section{Evaluation Setting}
\subsection{Hypotheses to Verify}
BERT's pre-training learns to generate sentence representations broadly transferable to different NLP tasks. 
In contrast, our method gives more direct supervision to generate representations suitable for semantic equivalence assessment tasks. 
We set up the following hypotheses on features of our method, which will be empirically verified through evaluation: 
{
\setlength{\leftmargini}{20pt}
\begin{itemize}
\setlength{\itemsep}{0pt}
\item[H1] Our method contributes to semantic equivalence assessment tasks.
\item[H2] Our method achieves improvement on downstream tasks that only have small amounts of training datasets for fine-tuning.
\item[H3] Our method moderately improves tasks if they are relevant to semantic equivalence assessment. 
\item[H4] Our training does not transfer to distant downstream tasks that are independent to semantic equivalence assessment.
\item[H5] Phrasal and sentential paraphrase classification complementarily benefits sentence representation learning.
\end{itemize}
}

\subsection{GLUE Datasets}
\begin{table}[!t]
\centering
\resizebox{\linewidth}{!}{%
\begin{tabular}{lll}
\hline
Corpus     & Task                             & Metrics  \\\hline
MRPC     & paraphrase                  & F$1$  \\
STS-B      & STS                             & Pearson corr.  \\
QQP       & paraphrase                 & F$1$  \\\hline
MNLI-m & in-domain NLI      & accuracy  \\
MNLI-mm & cross-domain NLI      & accuracy \\
RTE         & NLI                            & accuracy  \\
QNLI       & QA/NLI                             & accuracy  \\\hline
SST         & sentiment          & accuracy   \\
CoLA       & acceptability  & Matthews corr. \\\hline
\end{tabular}
}
\caption{GLUE tasks and evaluation metrics.}
\label{tb:glue}
\end{table}

\begin{table*}[!t]
\centering
\begin{tabular}{c|ccc|ccc|cc}
\hline
\multirow{2}{*}{\diagbox{Model}{Task}}                    &\multicolumn{3}{c|}{Semantic Equivalence}        & \multicolumn{3}{c|}{NLI}                                                             &\multicolumn{2}{c}{Single-Sent.} \\
                    & MRPC                     &STS-B                           & QQP              &MNLI (m/mm)                  & RTE                      &QNLI                       & SST              & CoLA                   \\\hline
BERT-base     & $88.3$                     & $84.7$                         &$71.2$            & $84.3/83.0$                    & $59.8$                  &$89.1$                    &$93.3$            &$52.7$           \\
BERT-large    & $\underline{88.6}$     & $\underline{86.0}$        &${\bf 72.1}$     & ${\bf 86.2}/{\bf 85.5}$      & ${\bf 65.5}$           &${\bf 92.7}$             &${\bf 94.1}$    &${\bf 55.7}$   \\\hline
Transfer Fine-Tuning     & ${\bf 89.2}$              & ${\bf 87.4}$                  &$71.2$            &$83.9/\underline{83.1}$    & $\underline{64.8}$  &$\underline{89.3}$   &$93.1$            &$47.2$    \\\hline
\end{tabular}
\caption{GLUE test results scored by the GLUE evaluation server. The best scores are represented in \textbf{bold} and scores higher than those of BERT-base are \underline{underlined}.}
\label{tb:overall_results}
\end{table*}

We empirically verified the hypotheses H1 to H5 using the General Language Understanding Evaluation (GLUE) benchmark~\cite{glue}\footnote{\url{https://gluebenchmark.com/}}, which is the standard benchmark and provides collections of datasets for natural language understanding tasks. 
\Tref{tb:glue} summarizes the tasks and evaluation metrics at GLUE. 
All the scores reported in this paper are computed at the GLUE evaluation server unless stated otherwise.
Accuracies on MRPC and QQP and Spearman correlation on STS-B are omitted due to space limitations. 
Note that they showed the same trends as F$1$ and Pearson correlation, respectively, in our experiment.
WNLI was excluded because the GLUE web site reports its issues.\footnote{\url{https://gluebenchmark.com/faq}}

GLUE tasks can be categorized according to their aims as follows.

\paragraph{Semantic Equivalence Assessment Tasks (MRPC, STS-B, QQP)}
These are the primary targets of our method, which are used to verify hypothesis H1. 
Paraphrase identification assesses semantic equivalence in a sentence pair by binary judgments. 
Microsoft Paraphrase Corpus (MRPC)~\cite{dolan-quirk-brockett:2004:COLING} consists of sentence pairs drawn from news articles, while Quora Question Pairs (QQP)\footnote{\url{https://data.quora.com/First-Quora-Dataset-Release-Question-Pairs}} consists of question pairs from the community QA website.

STS assesses semantic equivalence by grading. 
STS benchmark (STS-B)~\cite{cer-etal-2017-semeval} provides sentence pairs drawn from heterogeneous sources, which are human-annotated with a level of equivalence from $1$ to $5$.   

\paragraph{NLI Tasks (MNLI-m/mm, RTE, QNLI)}
We use natural language inference (NLI) tasks to verify hypothesis H3 because they constitute a class of problems relevant to semantic equivalence assessment. 
NLI tasks are different from semantic equivalence assessment in that they often require logical inference and understanding of common-sense knowledge.  
The Multi-Genre Natural Language Inference Corpus (MNLI)~\cite{williams-etal-2018-broad} is a crowd-sourced corpus and covers heterogeneous domains. 
MNLI-m is an in-domain NLI task while MNLI-mm is a cross-domain NLI task. 
The Recognizing Textual Entailment (RTE) corpus\footnote{\url{https://aclweb.org/aclwiki/Recognizing_Textual_Entailment}} was created from news and Wikipedia. 
Question-answering NLI (QNLI) was created from The Stanford Question Answering Dataset~\cite{rajpurkar-etal-2016-squad} on which all the sentences were drawn from Wikipedia. 


\paragraph{Single-Sentence Tasks (SST, CoLA)}
We use these tasks to verify hypothesis H4. 
They aim to estimate features in a single sentence, which has little interaction with semantic equivalence assessment in a sentence pair.  
The Stanford Sentiment Treebank (SST)~\cite{socher-EtAl:2013:EMNLP} task is a binary sentiment classification, while The Corpus of Linguistic Acceptability (CoLA)~\cite{warstadt2018neural} task is a binary classification of grammatical acceptability.  

\subsection{Fine-Tuning on Downstream Tasks}
\label{sec:fine-tuning-setting}
Once trained, our model can be used in exactly the same manner as the pre-trained BERT models. 
For fine-tuning our models and replicating BERT's results under the same setting,  we set the hyper-parameter values to those recommended in~\cite{bert}: a batch size of $32$, a learning rate of $3\mathrm{e}-5$, the number of training epochs to $4$, and a dropout probability of $0.1$.
We fine-tuned all the models on downstream tasks using the script provided in the Pytorch version of BERT.\footnote{{\tt run$\_$classifier.py} in \url{https://github.com/huggingface/pytorch-pretrained-BERT}}
For STS-B, we modified the script slightly to conduct regression instead of classification. 
All other hyper-parameters were set to the default values defined in the BERT's fine-tuning script.

For fair comparison, we kept the same hyper-parameter settings described above across all tasks and models. 
\newcite{phang-arxiv-2018} discussed that BERT performances become unstable when a training dataset with fine-tuning is small. 
In our evaluation, performances were stable when setting the same hyper-parameters, but further investigation is our future work. 

\section{Results and Discussion}
\subsection{Effect on Semantic Equivalence Assessment Tasks}
\label{sec:overall_results}

\Tref{tb:overall_results} shows fine-tuning results on GLUE; our model, denoted as Transfer Fine-Tuning, is compared against BERT-base and BERT-large. 
The first set of columns shows the results of semantic equivalence assessment tasks. 
Our model outperformed BERT-base on MRPC ($+0.9$ points) and STS-B ($+2.7$ points). 
Furthermore, it outperformed even BERT-large by $0.6$ points on MRPC and by $1.4$ points on STS-B, despite BERT-large having $3.1$ times more parameters than our model. 
\newcite{bert} described that the next-sentence prediction task in BERT's pre-training aims to train a model that understands sentence relations. 
Herein, we argue that such relations are effective at generating representations broadly transferable to various NLP tasks, but are too generic to generate representations for semantic equivalence assessment tasks. 
Our method allows semantic relations between sentences and phrases that are directly useful for this class of tasks to be learned.

These results support hypothesis H1, indicating that our approach is more effective than blindly enlarging the model size. 
A smaller model size is desirable for practical applications. 
We have also applied our method on the BERT-large model, but its performance was not much improved to warrant the larger model size.
Further investigation regarding pre-trained model sizes is our future work. 

\subsection{Effect of the Amount of Fine-Tuning Datasets}
\label{sec:train_size}
\begin{table}[!t]
\centering
\resizebox{\linewidth}{!}{%
\begin{tabular}{cc|cp{0.25\linewidth}}
\hline
Task                                  & Train. size & BERT-base & Transfer Fine-Tuning       \\\hline
\multirow{2}{*}{MRPC}    &$1$k              & $81.6$  &$88.1$ ($+6.5$) \\
                                         &all ($3.7$k)    &$89.4$   &$90.2$ ($+0.8$) \\\hline
 \multirow{2}{*}{STS-B}   &$1$k              &$83.4$   &$86.2$ ($+2.8$) \\
                                         &all ($5.7$k)    &$88.1$   &$90.1$ ($+2.0$) \\\hline
 \multirow{3}{*}{QQP}   &$1$k              &$69.9$  &$71.4$ ($+1.5$) \\
                                         &$5$k              &$75.5$   &$76.3$ ($+0.8$) \\
                                        &$10$k              &$77.0$   &$77.6$ ($+0.6$) \\
                                         &$20$k              &$79.6$   &$79.5$ ($-0.1$) \\
                                         &all ($364$k)   &$87.7$   &$87.7$ ($\pm 0.0$) \\\hline                                 
\end{tabular}
}
\caption{Development set scores of the BERT-base model and our model (and their differences) that were fine-tuned using subsamples and full-size training sets.}
\label{tb:training_size_effect}
\end{table}

Our method did not improve upon BERT-base for QQP. 
We consider this is because a large QQP training set ($364$k sentence pairs) allows the BERT model to converge to a certain optimum. 
This also relates to hypothesis H2.

To investigate the effect of the sizes of training sets, we fine-tuned our model and BERT-base for semantic equivalence assessment tasks using randomly subsampled training sets.  
\Tref{tb:training_size_effect} shows scores on the development sets.\footnote{We used the development set because the GLUE server allows only two submissions per day. Note that the number of training epochs for fine-tuning is fixed in our experiments, hence, the development set was not used for other purposes.}
The result clearly indicates that our method is more beneficial when a training dataset is limited on a downstream task, which supports hypothesis H2. 
This property is preferable for a transfer learning scenario that unsupervised sentence representation learning assumes.

Another factor that may affect the performance is domain mismatch between our paraphrase corpora and QQP corpus. 
The former was mostly collected from news while the latter was extracted from a social QA forum. 
In the future, we will investigate the effects of domains by generating multi-domain paraphrase corpora using a method proposed by~\newcite{P18-1042}.

\begin{table*}[!t]
\centering
\begin{tabular}{l|ccc|ccc|cc}
\hline
\multirow{2}{*}{\diagbox{Model}{Task}}                    &\multicolumn{3}{c|}{Semantic Equivalence}        & \multicolumn{3}{c|}{NLI}                                                             &\multicolumn{2}{c}{Single-Sent.} \\
                                               & MRPC                    &STS-B                     & QQP          &MNLI (m/mm)                & RTE                         &QNLI                     & SST               & CoLA    \\\hline
Transfer Fine-Tuning               & ${\bf 89.2}$           & $\underline{87.4}$  &${\bf 71.2}$         &$83.9/{\bf 83.1}$            &$\underline{64.8}$     &$\underline{89.3}$  &$93.1$            &$47.2$    \\\hline
BERT-base                                & $88.3$                  & $84.7$                   &${\bf 71.2}$  &${\bf 84.3}/83.0$            &$59.8$                     &$89.1$                   &${\bf 93.3}$     &${\bf 52.7}$    \\
~$+$sentence                          &$88.2$                  &${\bf 87.6}$              &$71.1$         &$83.2/82.8$                   &${\bf 66.2}$              &${\bf 90.2}$            &$92.4$            &$39.8$     \\
~$+3$way-PP                          &$88.2$                    &$\underline{85.8}$   &$70.9$         &$82.9/81.9$                    &$\underline{65.8}$    &$88.0$                    &$91.3$             &$32.6$     \\
~$+$binary-PP                        &$87.7$                    &$82.8$                    &$70.7$         &$83.7/82.2$                    &$61.2$                     &$87.6$                    &$92.5$             &$42.1$     \\\hline
\end{tabular}
\caption{Results of the ablation study where the best scores are represented in \textbf{bold} and scores higher than those of BERT-base are \underline{underlined}. The last three rows show performances when conducting only sentential paraphrase classification ($+$sentence), phrasal paraphrase classification ($+3$way-PP), and binary classification of phrasal paraphrase ($+$binary-PP), respectively.}
\label{tb:ablation_study}
\end{table*}

\subsection{Effect on NLI Tasks}
\label{sec:nli}
The second set of columns in \Tref{tb:overall_results} shows the results on NLI tasks. 
Our model presents moderate improvements on most NLI tasks, which supports hypothesis H3. 
We consider this is because the majority of NLI tasks that require inferences in one-direction, contrary to bi-directional entailment relations of paraphrases, are uni-directional.

Another reason is that our elaborate feature generation for the phrasal paraphrase classifier tightly fits the model for paraphrase identification. 
This contributes to performance improvements on this task, but sacrifices the model's generality on relevant tasks. 
We tackle this issue in our follow-up study reported in the Appendix.

Among NLI tasks, our model largely outperformed BERT-base by $5.0$ point on RTE.
This may be again due to the property of our method that brings improvement on tasks with a limited training set as RTE has only $2.5$k training sentence pairs.

\subsection{Effect on Single-Sentence Tasks}
The last two columns of \Tref{tb:overall_results} show results on single-sentence tasks; SST and CoLA, which are the most distant tasks from paraphrase classification. 
Our model presents a slightly lower score on SST compared to BERT-base and performed poorly on CoLA. 

One potential reason for this degradation is that our training takes a sentence pair as input, which may weaken the ability to model a single sentence. 
Another cause is attributable to similarities between our training and fine-tuning tasks. 
For SST, sentiment analysis could be adversarial toward paraphrase discrimination tasks. 
Although paraphrasal sentences tend to have the same sentiments, sentences with the same sentiments do not generally hold paraphrastic relations.
For CoLA, semantic relations unlikely contribute to determining grammatical acceptability, as required by CoLA task.

Together with the results in \Sref{sec:nli}, hypothesis H4 is supported; the effectiveness of our method depends on relevance between paraphrase discrimination and downstream tasks.  
Our future work will be to examine what characteristics of NLP tasks make our method less effective.

\subsection{Ablation Study}
To verify hypothesis H5, we conducted an ablation study that investigates independent effects of sentential and phrasal paraphrase classification.
\Tref{tb:ablation_study} shows the results; the last three rows show performances when conducting only sentential paraphrase classification, phrasal paraphrase classification, and binary classification of {\tt paraphrase} and {\tt in-paraphrase} pairs, respectively.
All the models were fine-tuned in the same manner as described in \Sref{sec:fine-tuning-setting}.

First, the results support the hypothesis; sentential and phrasal paraphrase classification complements each other on sentence representation learning. 
Our model achieved its best scores on MRPC, MNLI-m/mm, SST, and CoLA tasks by conducting both sentential and phrasal paraphrase classification simultaneously. 
Interestingly, these scores are higher than those when sentential and phrasal paraphrase classification are conducted independently. 
This is reasonable considering the process of fine-tuning. 
Sentential paraphrase classification directly affects the representation of {\tt [CLS]}, which is the primary tuning factor in fine-tuning for downstream tasks. 
Alternatively, phrasal paraphrase classification affects representations of phrases, which are the basis for generating the {\tt [CLS]} representation. 
Simultaneously conducting both sentential and phrasal paraphrase classification thus creates synergy. 

It is also obvious that the three-way classification of phrasal paraphrases, on which the model discriminates paraphrases, random combinations of phrases from a random pair of sentences, and random combinations of phrases in a paraphrasal sentence pair, is superior to binary classification.  
This shows that discriminating random combinations of phrases, which is a simpler and easier task, also contributes to representation learning.



\section{Conclusion}
We empirically demonstrate that sentential and phrasal paraphrase relations help sentence representation learning. 
While BERT's pre-training aims to generate generic representations transferable to a broad range of NLP tasks, our method generates representations suitable for the class of semantic equivalence assessment tasks. 
Our method achieves performance gains while maintaining the model size. 
Furthermore, it exhibits improvement on downstream tasks with limited amounts of training datasets for fine-tuning, which is a property crucial for transfer learning. 

In the future, we plan to investigate the effects of our method on different sizes of BERT models. 
Additionally, we will apply our model to improve the alignment quality of the phrase alignment model. 

\section*{Acknowledgments}
We appreciate the anonymous reviewers for their insightful comments and suggestions to improve the paper. 
This work was supported by JST, ACT-I, Grant Number JPMJPR16U2, Japan.

\bibliography{./paraphrase_alignment}
\bibliographystyle{acl_natbib}

\appendix
\section*{Appendix: Transfer Fine-Tuning with Simple Features}
\label{sec:appendix}
To further investigate effects of transfer fine-tuning using paraphrase relations on BERT, we designed a model that generates a simplest feature to input into the classifier in \Fref{fig:overview}.  
We assume that this method transmits learning signals to the underlying BERT in a more effective manner. 
Specifically, we use mean-pooling to generate representations of source and target phrases in \Eref{eq:max-pooling-s} and \Eref{eq:max-pooling-t}, respectively. 
These representations are simply concatenated as a feature representation and then fed into the classifier. 

\begin{table*}[!t]
\centering
\begin{tabular}{l|ccc|ccc}
\hline
\multirow{2}{*}{\diagbox{Model}{Task}}                    &\multicolumn{3}{c|}{Semantic Equivalence}        & \multicolumn{3}{c}{NLI}                                                             \\
                                                     & MRPC                         &STS-B                            & QQP                          &MNLI (m/mm)                                          & RTE                            &QNLI                           \\\hline 
BERT-base                                   & $88.3$                        & $84.7$                         &$71.2$                        & $84.3/83.0$                                            & $59.8$                       &$89.1$                        \\ 
Transfer Fine-tuning                  & $\underline{89.2}$   & $\underline{87.4}$     &$71.2$                        &$83.9/\underline{83.1}$                        &$\underline{64.8}$    &$\underline{89.3}$    \\ 
Simple Transfer Fine-Tuning    & $\underline{88.6}$   & ${\bf 87.7}$                 &$\underline{71.5}$   &$\underline{84.7}/\underline{83.6}$    & $\underline{67.0}$   &$\underline{91.1}$    \\\hline\hline 
BERT-large                                   & $88.6$                       & $86.0$                         &$72.1$                       & $86.2/85.5$                                             & $65.5$                       &${\bf 92.7}$                \\ 
Simple Transfer Fine-Tuning     & ${\bf 89.9}$               & $\underline{87.1}$     &${\bf 72.5}$              &${\bf 86.5}/{\bf 85.6}$                             &$\bf{68.2}$                &$92.2$                         \\\hline 
\end{tabular}
\caption{Test results on semantic equivalence assessment and NLI tasks scored by the GLUE evaluation server. The best scores for each task are represented in \textbf{bold}. The scores higher than those of BERT counterparts (against BERT-base and BERT-large, respectively) are \underline{underlined}. Our models with simple feature generation (Simple Transfer Fine-Tuning) consistently outperformed the BERT models and achieved the best scores for six out of seven tasks.}
\label{tb:overall_results_simplemodel}
\end{table*}

\begin{table*}[!t]
\centering
\begin{tabular}{l|ccc|ccc}
\hline
\multirow{2}{*}{\diagbox{Model}{Task}}                    &\multicolumn{3}{c|}{Semantic Equivalence}        & \multicolumn{3}{c}{NLI}                                                           \\ 
                                                  & MRPC                              &STS-B                             & QQP                           &MNLI (m/mm)                               & RTE                                 &QNLI                       \\\hline 
Simple Transfer Fine-Tuning  & $\underline{88.6}$          & ${\bf 87.7}$                  &${\bf 71.5}$                 &${\bf 84.7}/{\bf 83.6}$                  & ${\bf 67.0}$                    &${\bf 91.1}$       \\\hline 
BERT-base                                & $88.3$                              & $84.7$                           &$71.2$                         & $84.3/83.0$                                   & $59.8$                           &$89.1$                  \\ 
~$+$sentence                        &$88.2$                              &$\underline{87.6}$          &$71.1$                         &$83.2/82.8$                                     &$\underline{66.2}$         &$\underline{90.2}$ \\ 
~$+3$way-PP [Elaborate Feature]  &$88.2$                            &$\underline{85.8}$                  &$70.9$                          &$82.9/81.9$                                    &$\underline{65.8}$        &$88.0$                    \\ 
~$+3$way-PP [Simple Feature]       &${\bf 89.0}$                    &$\underline{86.6}$                   &${\bf 71.5}$                 &${\bf 84.7}/{\bf 83.6}$               &$\underline{65.6}$        &$\underline{90.6}$  \\\hline 
\end{tabular}
\caption{Results of the ablation study where the best scores are represented in \textbf{bold} and scores higher than those of BERT-base are \underline{underlined}. 
The third row shows performances when conducting only sentential paraphrase classification ($+$sentence) and the fourth row shows those when conducting only phrasal paraphrase classification with elaborate feature generation ($+3$way-PP [Elaborate Feature]), as reported in \Tref{tb:ablation_study}. 
The last row shows performances when conducing phrasal paraphrase classification with simple feature generation ($+3$way-PP [Simple Feature]). 
Results indicate that sentential and phrasal paraphrase classification complementarily contributes to Simple Transfer Fine-Tuning modeling.}
\label{tb:ablation_study_simplemodel}
\end{table*}

\Tref{tb:overall_results_simplemodel} compares this new model (denoted as Simple Transfer Fine-Tuning) to BERT models as well as our model with the elaborate feature generation described in \Sref{sec:our-model} (denoted as Transfer Fine-Tuning) on semantic equivalent assessment and NLI tasks of GLUE benchmark. 
\Tref{tb:ablation_study_simplemodel} reports an ablation study.  
The results and findings are summarized as follows. 
\begin{itemize}
\item Our model with simple feature generation (Simple Transfer Fine-Tuning) on BERT-base outperformed BERT on both semantic equivalent assessment and NLI  tasks. Furthermore, it performed on-par against BERT-large on MRPC and outperformed it on STS-B and RTE, despite BERT-large having $3.1$ times more parameters than our model. 
\item The same trend was confirmed on the model trained on BERT-large, where our model outperformed BERT-large on all the tasks except QNLI. 
\item Simple Transfer Fine-Tuning also outperformed our model with elaborate feature generation (Transfer Fine-Tuning) on all semantic equivalent assessment and NLI tasks except MRPC. This result implies that elaborate feature generation tightly fits the model to paraphrase identification while sacrifices its generality to relevant tasks. Further investigation will be our future work.    
\item Sentential and phrasal paraphrase classification complements each other on sentence representation learning when using simple feature generation, as also confirmed when using the elaborate feature generation in \Tref{tb:ablation_study}. Simple Transfer Fine-Tuning achieved higher scores on STS-B, RTE, and QNLI tasks than models trained either with only sentential ($+$sentence) or phrasal paraphrase ($+3$way-PP [Simple Feature]) classification.
\item Simple feature generation improves the performance of the model trained with only phrasal paraphrase classification; $+3$way-PP [Simple Feature] outperformed $+3$way-PP [Elaborate Feature] on all tasks except RTE.
\end{itemize} 

\end{document}